\documentclass{article}
\usepackage[T1]{fontenc}
\usepackage{spconf}
\usepackage{graphicx}
\usepackage{subcaption}
\usepackage{amsmath}
\usepackage{amssymb}
\usepackage{amsfonts}
\usepackage{enumitem}
\usepackage{xcolor}
\usepackage{url}
\usepackage{array}
\usepackage{multirow}
\usepackage{booktabs}
\usepackage[libertine]{newtxmath}
\usepackage{tikz}

\newcommand\copyrighttext{%
  \footnotesize \textcopyright~2026 IEEE. Personal use of this material is permitted.
  Permission from IEEE must be obtained for all other uses, in any current or future
  media, including reprinting/republishing this material for advertising or promotional
  purposes, creating new collective works, for resale or redistribution to servers or
  lists, or reuse of any copyrighted component of this work in other works.}
\newcommand\ieeecopyrightnotice{%
  \begin{tikzpicture}[remember picture,overlay]
    \node[anchor=south,yshift=8pt] at (current page.south)
      {\parbox{\dimexpr\textwidth\relax}{\copyrighttext}};
  \end{tikzpicture}%
}

\newcommand{\R}{\mathbb{R}}

\def\calE{{\mathcal E}}
\def\calN{{\mathcal N}}
\def\calL{{\mathcal L}}
\def\calP{{\mathcal P}}
\def\calS{{\mathcal S}}

\def\blurm{{\mathsf{BM}}}

\newcommand{\xx}{\mathbf{x}}
\newcommand{\zz}{\mathbf{z}}

\newcommand{\imgW}{0.125\linewidth}
\newcommand{\cropW}{0.125\linewidth}
\newcommand{\best}[1]{\textbf{#1}}
\newcommand{\second}[1]{\underline{#1}}
\newcommand{\blue}[1]{\textcolor{blue}{#1}}

\title{Controllable blind deblurring with diffusion models}
\name{Imane Si Salah$^{\star\dagger}$, Emile Cribelier$^\star$, Thomas Veit$^\star$, Wolf Hauser$^\star$, Arthur Leclaire$^\dagger$}
\address{DxO Labs, France$^\star$ \\
LTCI, Télécom Paris, IP Paris, France$^\dagger$}

\begin{document}
%
\maketitle
\ieeecopyrightnotice
\begin{abstract}
Image acquisition  with a camera involves several degradations due to the optical system, sensor, or low-level processing steps.
We address blind deblurring in professional photography: we aim to invert unknown isotropic blur without knowledge of the degradation kernel.
For such inverse problems,where some high-frequency information is lost, it is challenging to use generative models to produce details that are both photo-realistic and faithful to the input.
We propose SuperSharpen, a diffusion-based blind deblurring method offering explicit control over restoration strength through a blur measure. 
We compare two conditioning strategies: a ControlNet-style adapter on a frozen backbone, and full finetuning of the diffusion prior. Our experiments show that finetuning achieves better fidelity with fewer hallucinated details. We validate our approach on synthetic and real-world blur, demonstrating improved perceptual quality and controllable restoration strength.

\end{abstract}

\section{Introduction}
\label{sec:intro}

Many physical and environmental factors during image capture induce various forms of blur, which translate into a loss of high-frequency information.
These include optical imperfections in lenses, motion or camera shake, and low-light conditions that require longer exposure times. 
In addition, some preprocessing steps, such as RAW denoising, may further smooth out textures and fine details, introducing additional blur.
In this work, we address mild, spatially invariant perceived blur, including both slight defocus and smoothing introduced by denoising/ISP processing, while excluding motion blur and non-uniform blur.

Image restoration is a widely studied problem, and existing solutions can be classified as blind and non-blind~\cite{zhang2022deep}.
Non-blind methods assume that the degradation model is known, while blind approaches must estimate both the degradation and the clean image.
Traditionally, image restoration has been formulated as an optimization problem balancing a data-fidelity term with a regularization term. 
With the emergence of deep learning, learning-based solutions arose where neural networks are trained to learn an end-to-end mapping from the degraded image space to the clean image space~\cite{tao2018scale,liang2021swinir}. However, these methods often underperform in real-world scenarios due to the distribution shift between real and training data~\cite{wei2022rethinking}. Moreover, by minimizing a pixel-wise loss, they tend to predict an average of all plausible solutions, resulting in overly smoothed outputs that lack the fine details required for our restoration task.

Generative models~\cite{goodfellow2014generative,ho2020denoising} have emerged as strong candidates for image restoration as they constitute powerful priors to model natural images. The main challenge 
is then to condition sampling with a degraded observation.
In this work, we focus on diffusion models (DMs)~\cite{ho2020denoising,rombach2022high}. DMs generate images with a probabilistic denoising process, which iteratively refines pure white noise into a clean image.
Diffusion-based image restoration can be broadly divided into two categories. 
The first one~\cite{choi2021ilvr, chung2022diffusion,zhu2023denoising}, based on a pretrained diffusion scheme, consists in guiding the iterative sampling process using a kind of data consistency to the degraded input. 
One major asset of these methods is to avoid task-specific retraining.
However, they necessitate many sampling steps to reinforce the fidelity to the degraded observation. This limits the use of accelerated diffusion models like consistency models~\cite{song2023consistency}.
The second category involves end-to-end retraining to condition the DM at each step by the low-quality input.
This is done either by 1)~training an adapter network like ControlNet-style~\cite{zhang2023adding} that will be plugged into a frozen pretrained prior~\cite{lin2024diffbir,yang2024pixel,wu2024seesr}, or 2)~by finetuning the diffusion denoiser to condition on the degraded input~\cite{saharia2022palette,whang2022deblurring, saharia2022image}. Although these approaches require additional training, they often reach higher performance in terms of perceptual quality.
In this work, we investigate both these conditioning strategies on a latent diffusion model~\cite{rombach2022high} and compare them in the context of blind image deblurring.  

In addition, most existing diffusion-based restoration methods do not provide an explicit control over the desired generative restoration level.
We address this limitation by introducing the ``blur measure'' as conditional input: at training time, it represents the estimated blur level of the used degradation, while at inference it represents the quantity of blur that one wishes to remove.

Our contributions can be summarized as follows:
\begin{itemize}[noitemsep,topsep=0pt,leftmargin=*]
\item We propose SuperSharpen (SupS), a blind deblurring model based on latent diffusion, which takes as input the degraded image and a so-called blur measure~($\blurm$), which allows for explicit control over the restoration strength at inference.
\item We empirically compare two conditioning methods of the diffusion backbone (ControlNet-based and finetuning) and show that finetuning achieves better fidelity to the degraded input with fewer hallucinated details.
\item We validate our method on both synthetic and real-world blur, including blur arising from optical imperfections and RAW processing.
\end{itemize}

\begin{figure}[!h]
  \centering

  \begin{subfigure}[t]{\linewidth}
    \centering
    \includegraphics[width=0.9\linewidth]{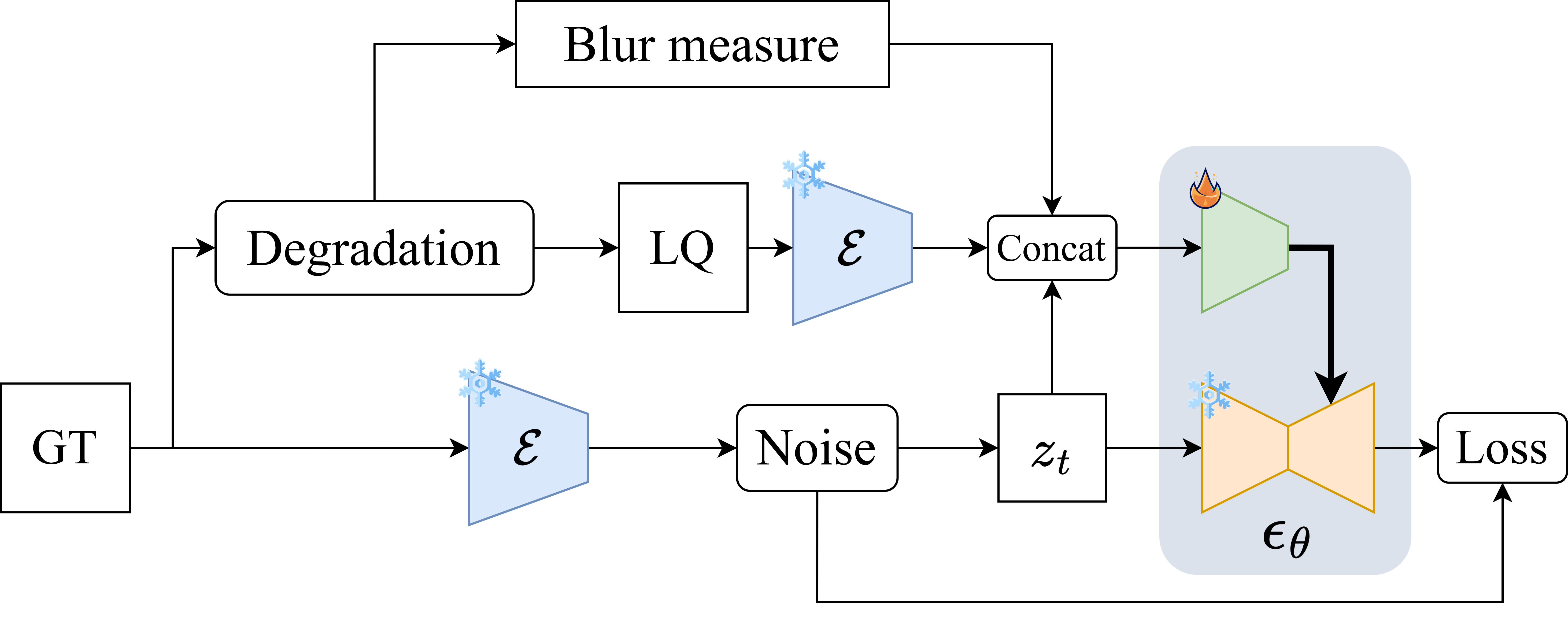}
    \caption{SupS-CN training setup}
    \label{fig:cldm_train}
  \end{subfigure}

  \vspace{0.1em}

  \begin{subfigure}[t]{\linewidth}
    \centering
    \includegraphics[width=0.9\linewidth]{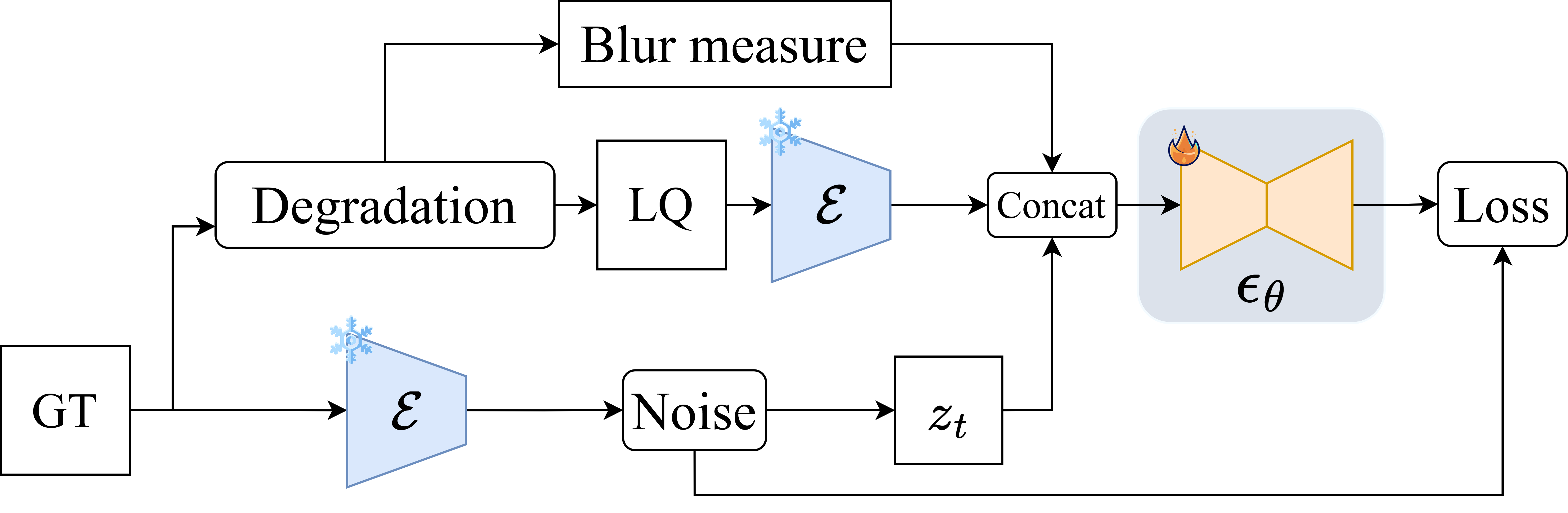}
    \caption{SupS-FT training setup}
    \label{fig:ldm_train}
  \end{subfigure}

  \vspace{-0.1em}

  \begin{subfigure}[t]{\linewidth}
    \centering
    \includegraphics[width=\linewidth]{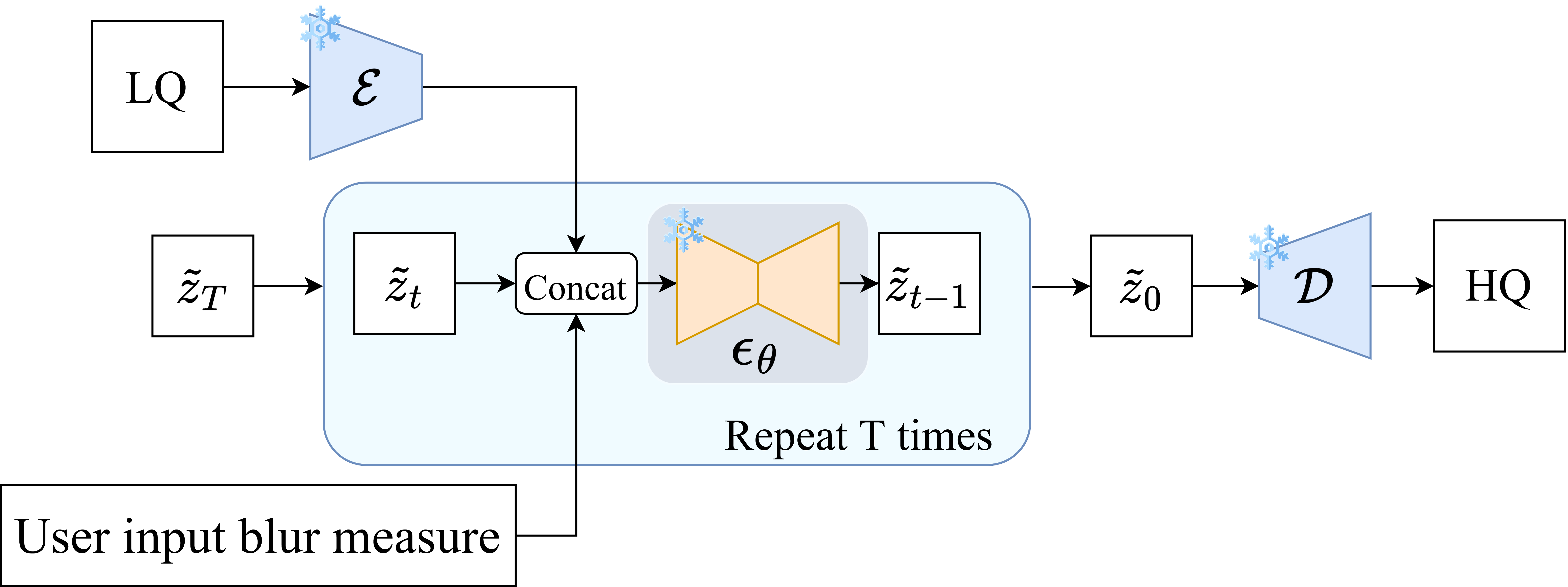}
    \caption{SupS-FT inference setup}
    \label{fig:ldm-sample}
  \end{subfigure}
  \caption{
  We present the training setups for (a) ControlNet-based conditioning (SupS-CN) and (b) finetuning-based conditioning (SupS-FT).
  Both models rely on a conditional score-based latent denoiser $\epsilon_\theta$ that takes as input the encoded Low Quality image and the corresponding blur measure in addition to the encoded Ground Truth to which we add noise $\zz_t$. In (a) only the ControlNet adapters are trained (\includegraphics[width=6pt]{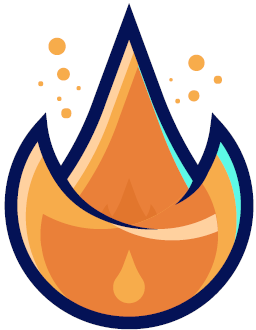}) while the diffusion backbone is frozen (\includegraphics[width=6pt]{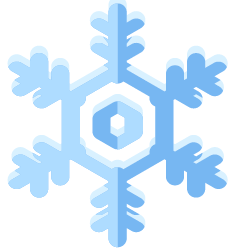}). In (b) the entire diffusion denoiser is finetuned~(\includegraphics[width=6pt]{train_symbol.png}).
   Noting that in both training setups, the VAE encoder $\calE$ is frozen (\includegraphics[width=6pt]{frozen_symbol.png}).
  In (c), we illustrate the inference setup only for SupS-FT which generates the high-quality image by iterative use of $\epsilon_\theta$ conditioned by the degraded image and a user-defined blur measure to control the restoration strength. Notice that at training $t$ is sampled randomly while at inference, $t$ is decreased from $T$ to $0$.
  }
  \label{fig:train_sample}
\end{figure}

\begin{figure}[t]
    \centering
    \includegraphics[width=\linewidth]{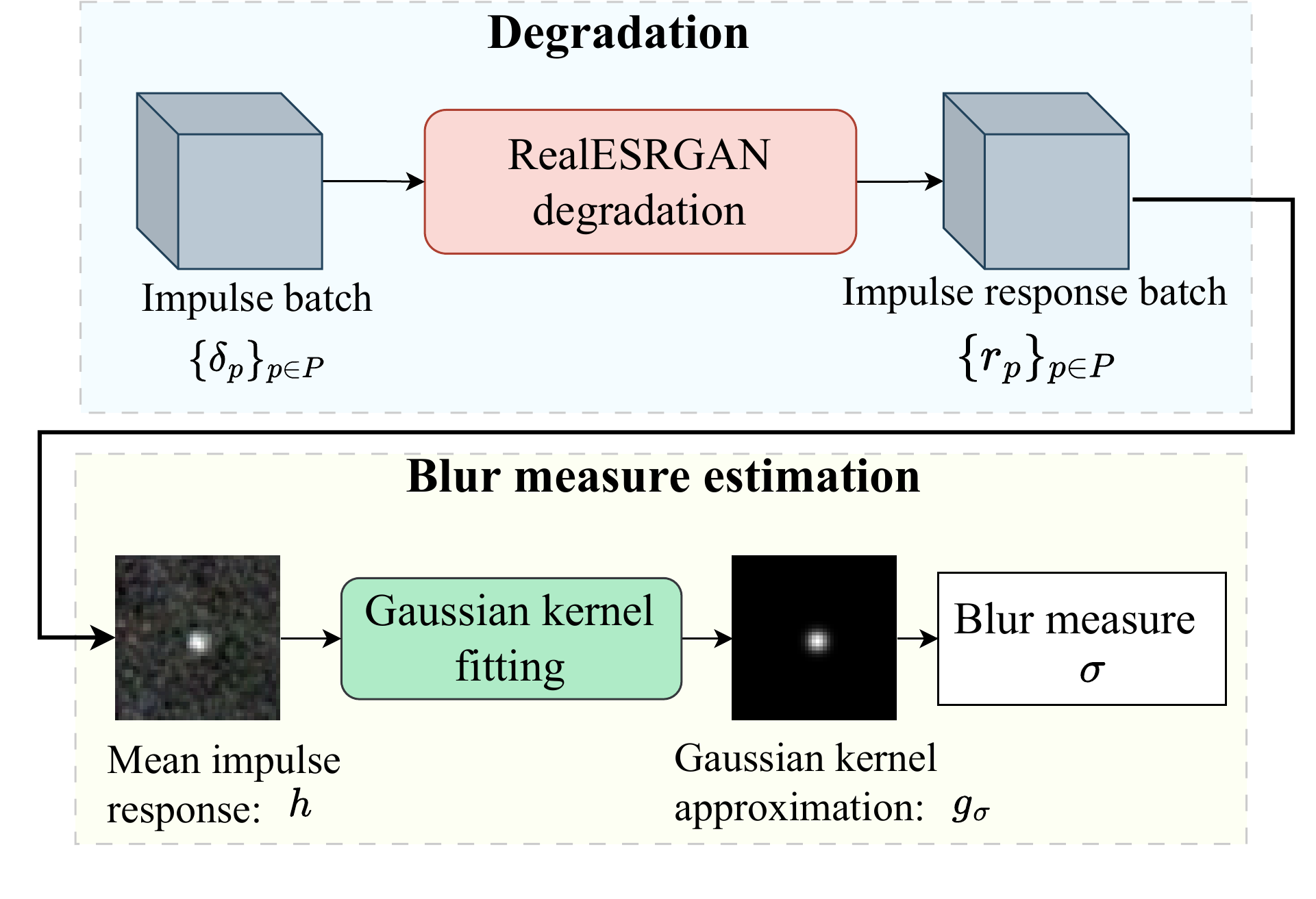} 
    \caption{Blur measure computation setup. We estimate the effective blur kernel of the degradation pipeline by computing its impulse response from multiple impulses located at different spatial locations. We then fit an isotropic Gaussian kernel to the average impulse response and define our blur measure as the standard deviation of the fitted Gaussian.}
    \label{fig:blur_measure_pipeline}
\end{figure}

\begin{figure*}[t]
\setlength{\tabcolsep}{1pt} 
\centering
\begin{tabular}{cccccccc}
{\small LQ} & {\small ResShift} & {\small PASD} & {\small SeeSR} & {\small DiffBIR} & {\small SupS-CN} & {\small SupS-FT} & {\small HQ} \\[0.1em]
{\small PSNR$\uparrow$/ LPIPS$\downarrow$} & {\small 19.45/0.20} & {\small 22.77/0.26} & {\small 20.06/0.40} & {\small 21.08/0.71} & {\small 18.88/0.27} & {\small 20.28/0.18} & {\small --} \\
\includegraphics[width=\imgW]{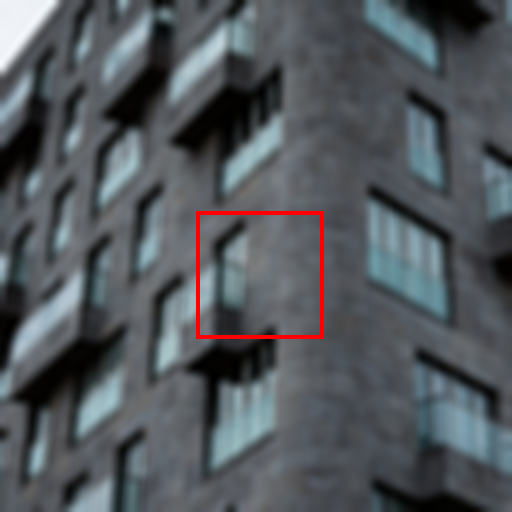} &
\includegraphics[width=\imgW]{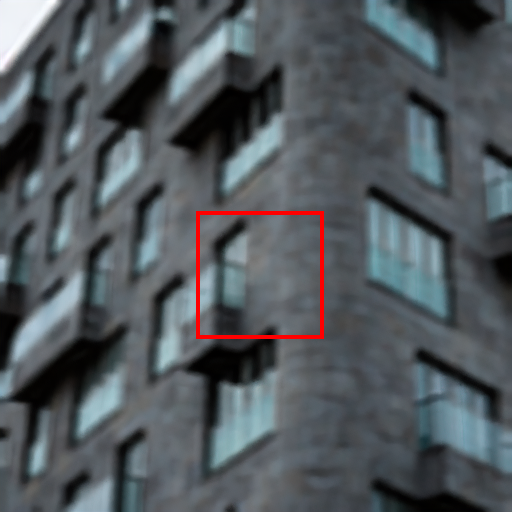} &
\includegraphics[width=\imgW]{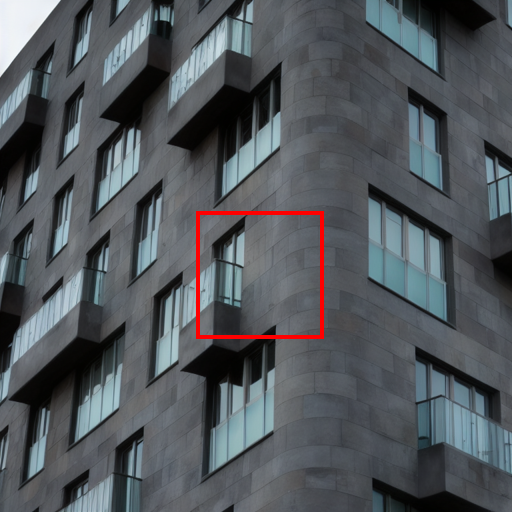} &
\includegraphics[width=\imgW]{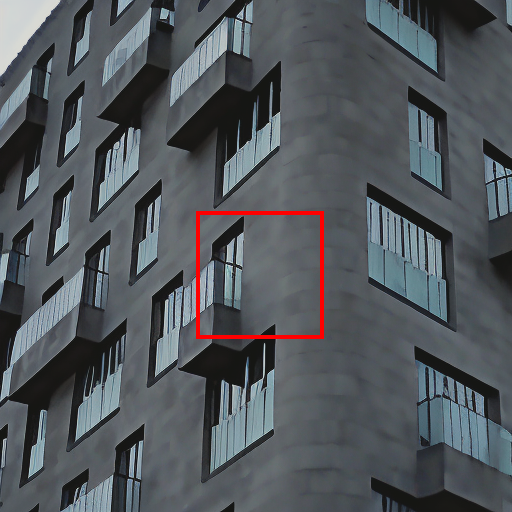} &
\includegraphics[width=\imgW]{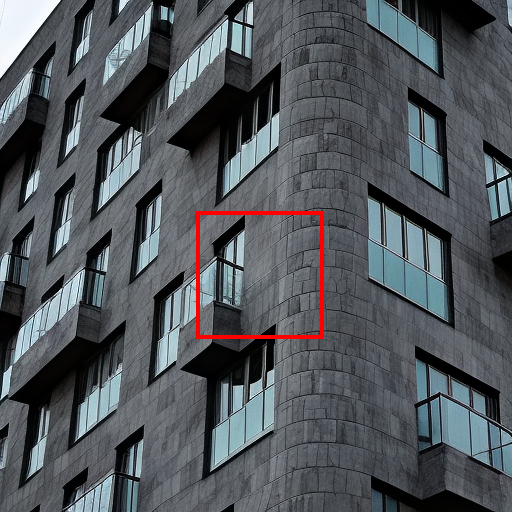} &
\includegraphics[width=\imgW]{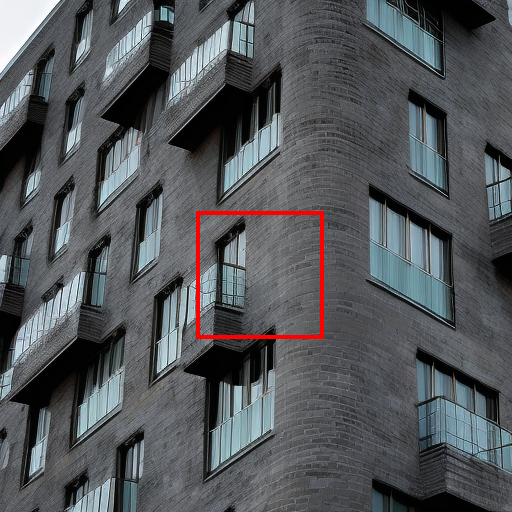} &
\includegraphics[width=\imgW]{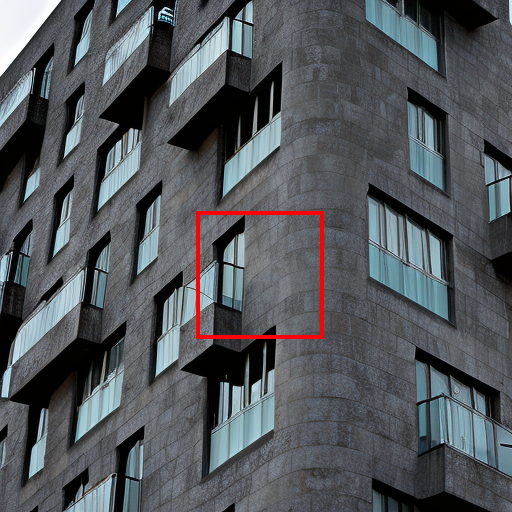} &
\includegraphics[width=\imgW]{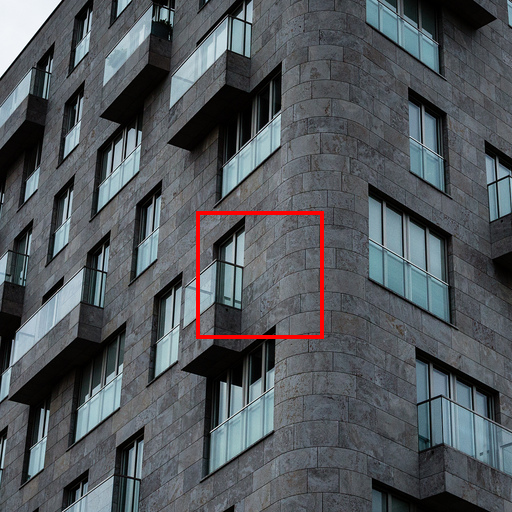} \\ [-0.3em]
\includegraphics[width=\cropW]{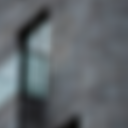} &
\includegraphics[width=\cropW]{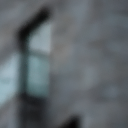} &
\includegraphics[width=\cropW]{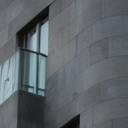} &
\includegraphics[width=\cropW]{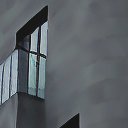} &
\includegraphics[width=\cropW]{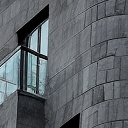} &
\includegraphics[width=\cropW]{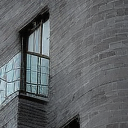} &
\includegraphics[width=\cropW]{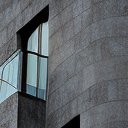} &
\includegraphics[width=\cropW]{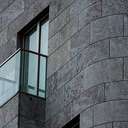} \\[0.1em]
{\small PSNR$\uparrow$/ LPIPS$\downarrow$} & {\small 19.13/0.08} & {\small 22.14/0.13} & {\small 20.50/0.14} & {\small 19.06/0.46} & {\small 18.03/0.13} & {\small 20.09/0.07} & {\small --} \\
\includegraphics[width=\imgW]{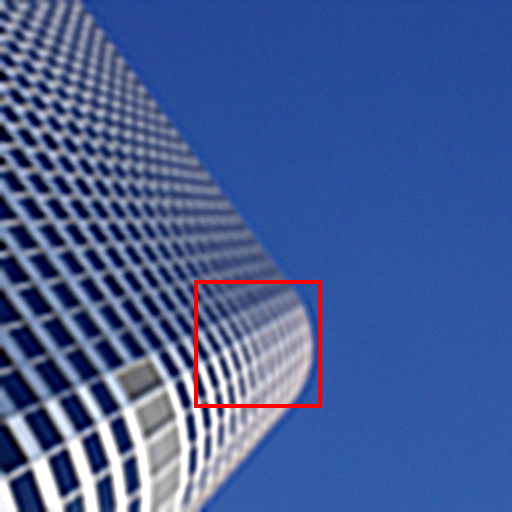} &
\includegraphics[width=\imgW]{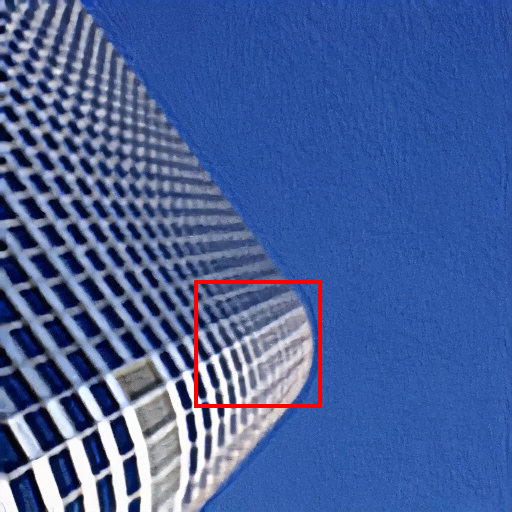} &
\includegraphics[width=\imgW]{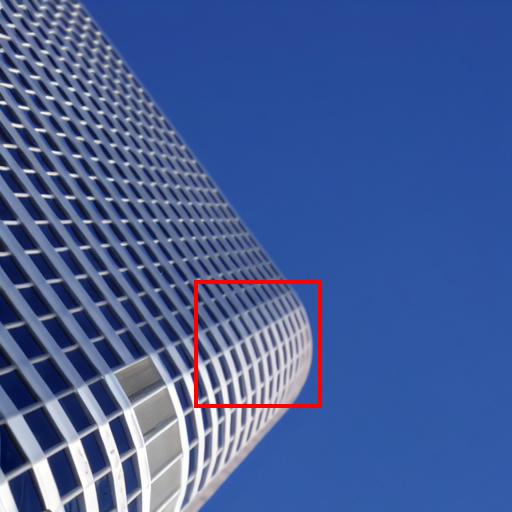} &
\includegraphics[width=\imgW]{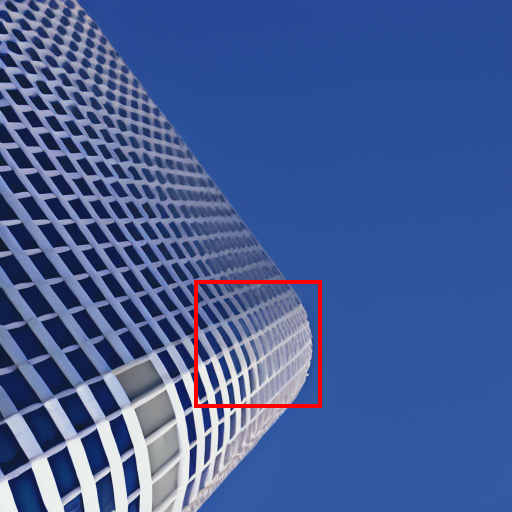} &
\includegraphics[width=\imgW]{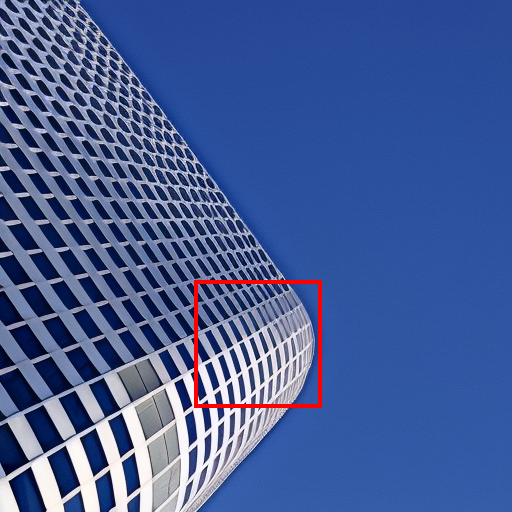} &
\includegraphics[width=\imgW]{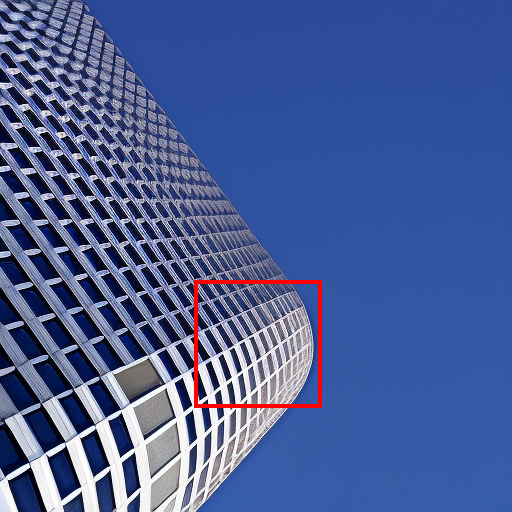} &
\includegraphics[width=\imgW]{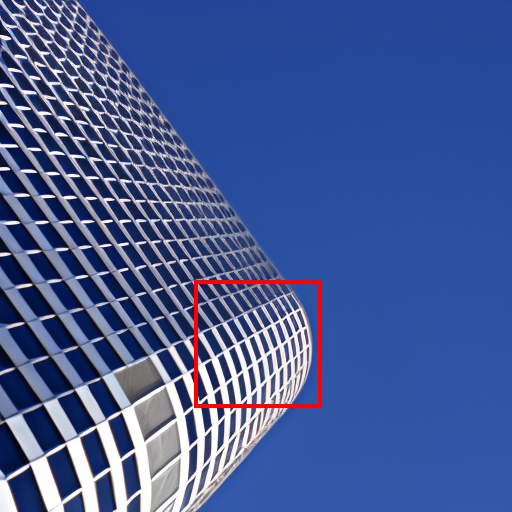} &
\includegraphics[width=\imgW]{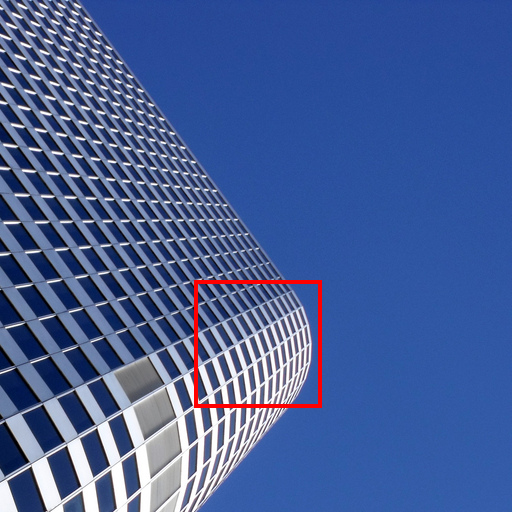} \\ [-0.3em]
\includegraphics[width=\cropW]{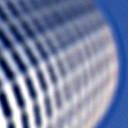} &
\includegraphics[width=\cropW]{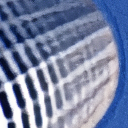} &
\includegraphics[width=\cropW]{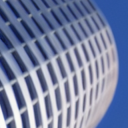} &
\includegraphics[width=\cropW]{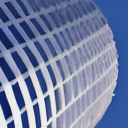} &
\includegraphics[width=\cropW]{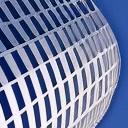} &
\includegraphics[width=\cropW]{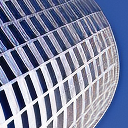} &
\includegraphics[width=\cropW]{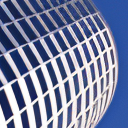} &
\includegraphics[width=\cropW]{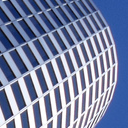} \\
\end{tabular}
\caption{Qualitative comparison on images with synthetic blur.
 The sharpest results are obtained with DiffBIR, SupS-CN and SupS-FT, the latter producing less unplausible structures.
}
\label{fig:qual_results}
\end{figure*}

\section{Blind Deblurring Network}
\label{sec:method}
\subsection{Preliminaries: Diffusion models}
\label{sec:diffusion_models}

Our method builds on a Stable Diffusion (SD)~\cite{rombach2022high} backbone, a latent diffusion model (LDM) that uses a variational autoencoder (VAE) to operate in a lower-dimensional latent space.
Let $\xx_0 \in \R^{H \times W \times 3}$ be a clean image and $\zz_0 = \calE(\xx_0) \in \R^{\frac{H}{s} \times \frac{W}{s} \times c}$ be its latent representation, where $\calE$ is the encoder, $s$ is the spatial downsampling factor, and $c$ is the number of latent channels.
During the forward diffusion process, Gaussian noise is progressively added to the latent code over $T$ timesteps. At timestep $t \in \{1,\ldots,T\}$, the noisy latent is given by $\zz_t = \sqrt{\bar{\alpha}_t} \, \zz_0 + \sqrt{\bar{\beta}_t} \, \boldsymbol{\epsilon}_t$, where $\boldsymbol{\epsilon}_t \sim \calN(\mathbf{0}, I_d)$ and $\bar{\alpha}_t$, $\bar{\beta}_t$ depend on a predefined noise schedule.
A neural network $\boldsymbol{\epsilon}_\theta(\zz_t, t)$ with parameters $\theta$ is trained to predict the noise $\boldsymbol{\epsilon}_t$ by minimizing
\begin{equation}
\calL(\theta)
= \mathbb{E}_{t,\,\zz_0,\,\boldsymbol{\epsilon}_t}
\left[ \,\big\|\boldsymbol{\epsilon}_t - \boldsymbol{\epsilon}_\theta(\zz_t, t)\big\|_2^2 \,\right] .
\label{eq:ddpm_loss}
\end{equation}
At inference, the reverse sampling process iteratively denoises from pure noise $\tilde{\zz}_T \sim \calN(\mathbf{0}, I_d)$ to generate a clean latent $\tilde{\zz}_0$, which is then decoded to produce the output image, see the details in~\cite{ho2020denoising}.

\subsection{Diffusion model conditioning}

While the DDPM formulation above describes an unconditional generative model, we address blind deblurring as generation conditioned by a blurry input image. 
This conditioning is learned using pairs of high-quality (HQ) images and their corresponding low-quality (LQ) images. During training, each LQ image is obtained by applying the same degradation pipeline as used in Real-ESRGAN~\cite{wang2021real}. 
Let $\zz_{LQ}= \calE(\xx_{LQ})$ denote the latent code of the degraded image and $\zz_t$ the noisy latent at time~$t$. We investigate two types of conditioning categories for injecting the degraded input into the diffusion model. 

The first category, SuperSharpen ControlNet (SupS-CN), uses ControlNet layers~\cite{zhang2023adding}.
As shown in Fig.~\ref{fig:cldm_train},
the ControlNet adapter is initialized as a trainable copy of the encoder and middle blocks of the diffusion denoiser U-Net, while the pretrained diffusion backbone is kept frozen. At each diffusion step, the conditioning branch takes the concatenation of $\zz_{LQ}$ and $\zz_t$, extracts joint features, and injects them into the frozen backbone through residual connections.
SupS-CN is inspired by DiffBIR~\cite{lin2024diffbir}, which uses a two-stage restoration model: a preprocessing module
followed by a ControlNet-based generative module. In comparison, our approach omits the preprocessing network. 

The second category, SuperSharpen Finetuned (SupS-FT), shown in Fig.~\ref{fig:ldm_train}, directly integrates the conditioning input into the diffusion backbone and therefore requires finetuning the entire diffusion denoiser. We extend the diffusion U-Net to accept the concatenation of $\zz_{LQ}$ and $\zz_t$.

In both settings, the models are trained with the standard noise-prediction objective~\eqref{eq:ddpm_loss}, extended to take the LQ latent as an additional input:
\begin{equation}
\calL(\theta)
= \mathbb{E}_{t,\,\zz_0,\,\boldsymbol{\epsilon}_t} 
\left[ \,\big\|\boldsymbol{\epsilon}_t - \boldsymbol{\epsilon}_\theta(\zz_t, t,\zz_{LQ})\big\|_2^2 \,\right].
\label{eq:conditional_loss}
\end{equation}

The inference is run with the backward latent diffusion process, with the $\zz_{LQ}$ being reinjected in $\boldsymbol{\epsilon}$ at each reverse diffusion step, as illustrated in Fig.~\ref{fig:ldm-sample}.

\begin{figure*}[t]
\centering
\setlength{\tabcolsep}{1pt} 
\begin{tabular}{ccccccc}
{\small LQ} & {\small ResShift} & {\small PASD} & {\small SeeSR} & {\small DiffBIR} & {\small SupS-CN} & {\small SupS-FT} \\[0.1em]
\includegraphics[width=0.14\linewidth]{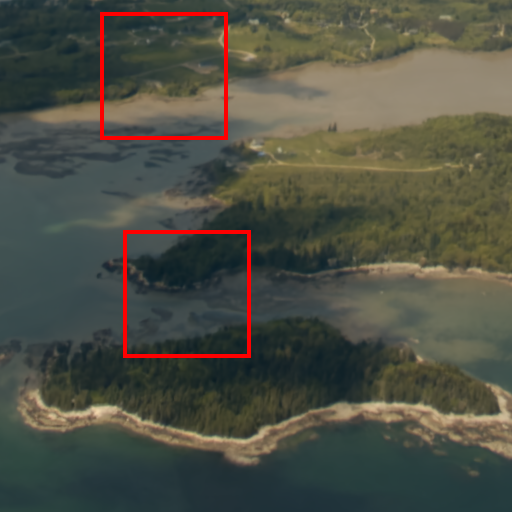} &
\includegraphics[width=0.14\linewidth]{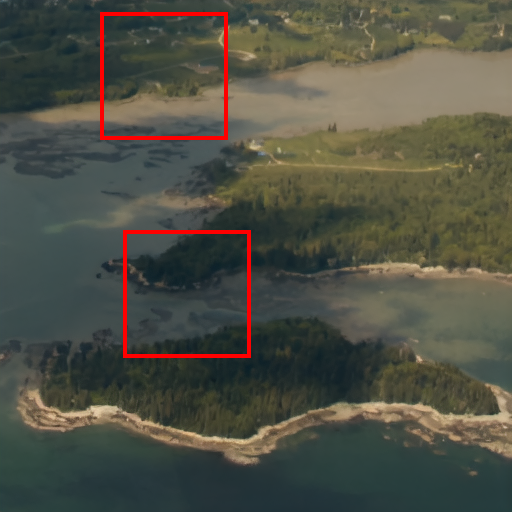} &
\includegraphics[width=0.14\linewidth]{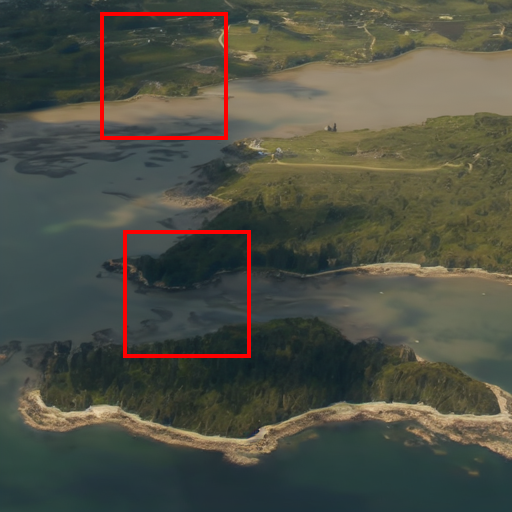} &
\includegraphics[width=0.14\linewidth]{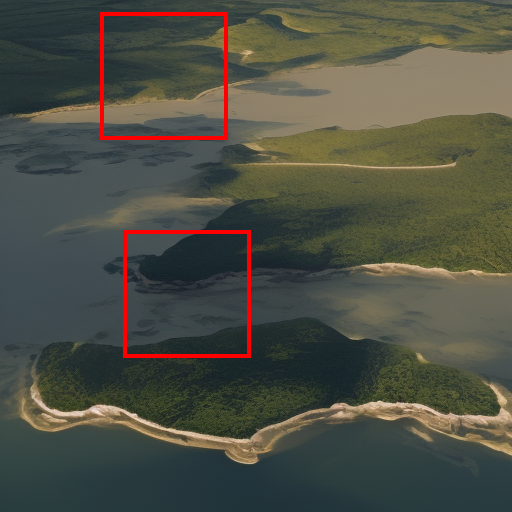} &
\includegraphics[width=0.14\linewidth]{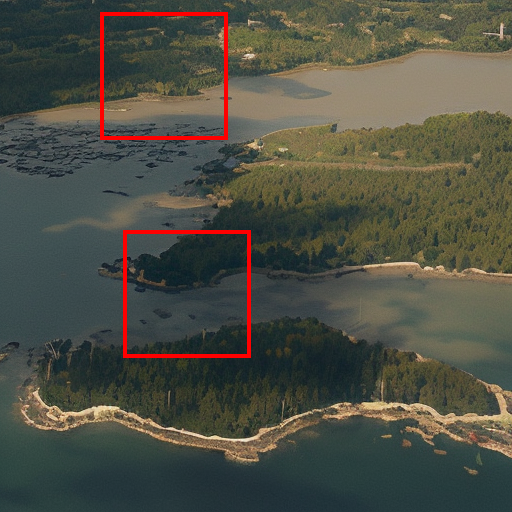} &
\includegraphics[width=0.14\linewidth]{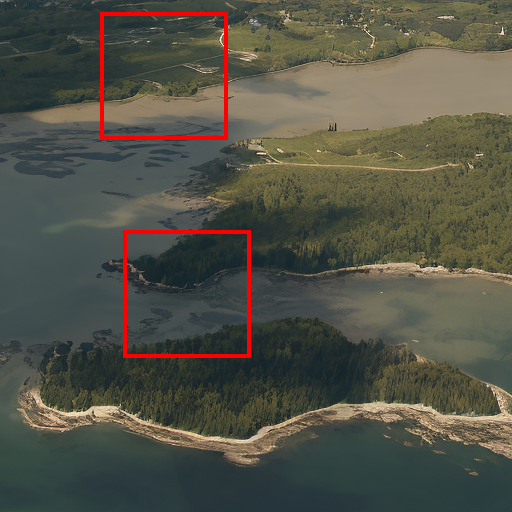} &
\includegraphics[width=0.14\linewidth]{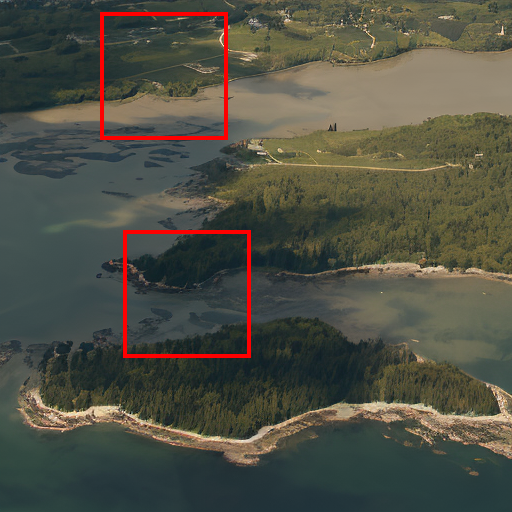} \\ [-0.2em]
\includegraphics[width=0.14\linewidth]{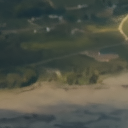} &
\includegraphics[width=0.14\linewidth]{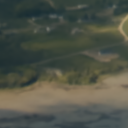} &
\includegraphics[width=0.14\linewidth]{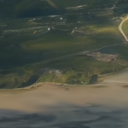} &
\includegraphics[width=0.14\linewidth]{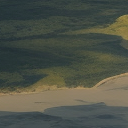} &
\includegraphics[width=0.14\linewidth]{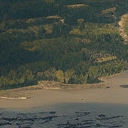} &
\includegraphics[width=0.14\linewidth]{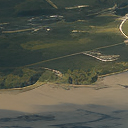} &
\includegraphics[width=0.14\linewidth]{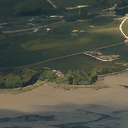} \\ [-0.2em]
\includegraphics[width=0.14\linewidth]{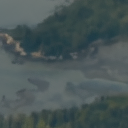} &
\includegraphics[width=0.14\linewidth]{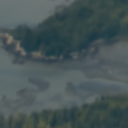} &
\includegraphics[width=0.14\linewidth]{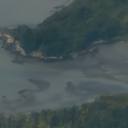} &
\includegraphics[width=0.14\linewidth]{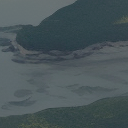} &
\includegraphics[width=0.14\linewidth]{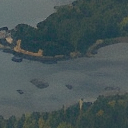} &
\includegraphics[width=0.14\linewidth]{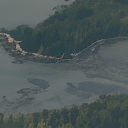} &
\includegraphics[width=0.14\linewidth]{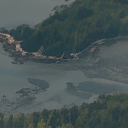} 

\end{tabular}
\caption{Qualitative comparison on real blurry images. Compared to the other methods, our SupS models attain a better compromise in terms of photo-realism and fidelity to the input.}
\label{fig:real_comparison}
\end{figure*}

\subsection{Blur measure estimation}
\label{subsec:blur_measure}

In addition to blind deblurring, we aim to provide explicit control over the restoration strength.
For that, we introduce a blur measure $\blurm\in\mathbb{R}_+$, used as an additional conditioning input.
During training, 
for each training sample, we compute the $\blurm$ value as an estimate of the blur introduced by the degradation pipeline (see below).
During inference on real images, $\blurm$ can be set by the user to adjust the restoration strength.

Our restoration model is trained with the degradation pipeline used to train \mbox{RealESRGAN}~\cite{wang2021real}, which combines various types of simulated degradations (isotropic and non-isotropic blur, resizing, noise addition, and JPEG compression), which is crucial to increase the model robustness to real-world degradations.
But since the degradation does not consist solely of blur, we propose to estimate the amount of blur it introduces by fitting an isotropic Gaussian blur kernel to the \emph{effective} impulse response of the degradation pipeline.
More precisely, given the degradation operator $A$, we average the responses $r_p = A(\delta_p)$ obtained at pre-defined positions $p \in \mathcal{P}$ taken on a $21 \times 21$ grid (because $A$ is not exactly translation-invariant).
For each $p$, the response is recentered using a shift $\calS_p$ computed with the center of mass of $r_p$.
We average over $p$ to get the estimated impulse response
$h = \frac{1}{|\calP|} \sum_{p \in \calP} \calS_p(r_p)$.
Finally, we fit an isotropic Gaussian kernel $g_{\sigma}$ to get the estimated blur measure of the degradation
\begin{equation}
\widehat{\blurm}(A) = \arg\min_{\sigma\in [\sigma_{\min},\sigma_{\max}]} \mathrm{MSE}\left(h, g_{\sigma}\right),
\end{equation}
This procedure is illustrated in Fig.~\ref{fig:blur_measure_pipeline}. While $\blurm$ is derived from a blur estimate during training, at inference, however, it does not need to match the exact physical blur present in the input image. Instead, it serves as a user-adjustable control signal: higher values lead to stronger detail generation, allowing users to tune the restoration strength according to their preference.

\section{Experiments}
\label{sec:experiments}
\textbf{Model architecture.}
We use a Stable Diffusion (SD2.1) model as our generative backbone. We evaluate two conditioning methods: (i) ControlNet-based conditioning: we freeze the SD2.1 U-Net and train only the ControlNet adapter layers that take the concatenation of the noisy and degraded latents in addition to the blur measure scalar extended to the latent dimensions. (ii) Backbone finetuning: we finetune the SD2.1 U-Net and extend its first layer to take the degraded input image and the blur measure as additional conditioning. 

\textbf{Training and data.}
We train on the LSDIR dataset~\cite{li2023lsdir}, where we generate paired training samples using a degradation pipeline similar to the one used for Real-ESRGAN~\cite{wang2021real} but with a scale factor of $1$ (since the objective is to restore images at the same resolution as the input) and with lower degradation parameters (because our network focuses on slight degradations).
In addition to synthesizing the LQ images, we make another pass through the degradation pipeline to compute the blur measure $\blurm$ as described in Section~\ref{subsec:blur_measure}. For that, the input impulses have resolution $75 \times 75$, and we use $|\calP|=441$ impulses located on the central $21 \times 21$ grid. The search interval for $\sigma$ is set to $[\sigma_{\min},\sigma_{\max}]=[0.1, 10.0]$. 
We input the scalar $\blurm$ value as a constant map concatenated with the latent $\zz_{LQ}$.

To make the model produce outputs with varying sharpness levels, we employ a data augmentation scheme where HQ images are convolved with a Gaussian kernel before degradation. Initially, this augmentation is applied to all training samples; then, as training progresses, to only $10\%$ of the dataset, forcing the model to learn to generate sharp outputs.

We train SupS-CN (resp. SupS-FT) for $270k$ (resp. $380k$) iterations using the AdamW optimizer with batches of $24$ crops of size $512 \times 512$. The learning rate is set to $10^{-4}$ and reduced to $10^{-5}$ for the last $100k$ iterations.
Training is performed on $4 \times$H100 GPUs during 13 days.
\begin{table}
\centering
\caption{Quantitative results on synthetically degraded images from Urban100 dataset. Best and second best results are highlighted in \textbf{bold} and \underline{underlined}, respectively.}
\label{tab:quant_results}
\scriptsize
\setlength{\tabcolsep}{3pt}
\resizebox{\linewidth}{!}{%
\begin{tabular}{lrrrrrr}
\toprule
& \multicolumn{3}{c}{Full-reference} & \multicolumn{3}{c}{No-reference} \\
\cmidrule(lr){2-4}\cmidrule(lr){5-7}
Method & PSNR $\uparrow$ & SSIM $\uparrow$ & LPIPS $\downarrow$ & MUSIQ $\uparrow$ & MANIQA $\uparrow$ & CLIP-IQA $\uparrow$ \\
\midrule
\blue{HQ }    & \blue{-- } & \blue{1.00} & \blue{0.00} & \blue{70.4433} & \blue{0.5534} & \blue{0.6762} \\
\midrule
DiffBIR          & 18.3051 & 0.5284 & 0.2278 & \second{71.6563} & 0.5986 & \second{0.7107} \\
PASD             & \best{21.9344} & \best{0.6518} & 0.2138 & 68.9798 & 0.4850 & 0.5904 \\
SeeSR            & 20.1731 & 0.5808 & 0.2463 & 70.4701 & 0.5529 & 0.6541 \\
ResShift         & 19.7569 & 0.4976 & 0.4138 & 48.0120 & 0.2960 & 0.4199 \\
\midrule
SupS-CN  & 19.7659 & 0.5549 & \second{0.2020} & \best{71.7406} & \best{0.6493} & \best{0.7299} \\
SupS-FT   & \second{20.7670} & \second{0.6138} & \best{0.1714} & 71.0819 & \second{0.6112} & 0.7102 \\
\bottomrule
\end{tabular}%
}
\end{table}

\textbf{Evaluation on synthetic data.}
We evaluate on images from Urban100~\cite{huang2015single}, synthetically degraded using a Gaussian blur kernel with std drawn in $[0.7, 7.5]$ and additive Gaussian noise with std drawn in $[0, 7]$. While we train with the Real-ESRGAN degradation pipeline to learn robustness across a broad degradation space, we test only with isotropic Gaussian blur,  which is a good single-parameter baseline.

We compare our method to several recent generative-based restoration methods: \mbox{DiffBIR}~\cite{lin2024diffbir}, PASD~\cite{yang2024pixel}, SeeSR \cite{wu2024seesr}, and ResShift~\cite{yue2023resshift}.
All these methods are based on DMs. \mbox{DiffBIR}, PASD, and SeeSR use a two-stage pipeline with a preprocessing module followed by a ControlNet-based generative module, while ResShift introduces a residual-shifting framework for efficient restoration.
Notice that these methods are designed for super-resolution but can be used with a scale factor of~$1$.
We report both full-reference metrics (PSNR, SSIM~\cite{wang2004image}, LPIPS~\cite{zhang2018unreasonable}) and no-reference metrics (MUSIQ~\cite{ke2021musiq}, MANIQA \cite{yang2022maniqa}, CLIP-IQA~\cite{wang2023exploring}).

Quantitative results are shown in Table~\ref{tab:quant_results}.
SupS-CN achieves the highest no-reference scores among all methods.
Interestingly, most methods even surpass the HQ for these no-reference metrics.
SupS-FT achieves the best LPIPS, indicating good perceptual fidelity.
However, it underperforms in terms of PSNR and SSIM compared to PASD, whose training includes a pixel-wise loss in its degradation removal module.
This suggests that finetuning the diffusion backbone allows for better fidelity to the degraded input, compared to ControlNet, since it is less dependent on a frozen prior.

Qualitative results in Fig.~\ref{fig:qual_results} confirm these observations: SupS-FT produces details that are more consistent with the HQ image, while SupS-CN produces images with similar sharpness level but less accurate textures.
This is why we favored SupS-FT over SupS-CN for the next assessment on real images.
From Fig.~\ref{fig:qual_results}, we also observe that the results obtained with our models and with \mbox{DiffBIR} are significantly better than those of the other methods. Moreover, SupS-FT introduces fewer high-frequency artifacts than \mbox{DiffBIR}.

\textbf{Evaluation on real data.}
We evaluate on images from Adobe5K~\cite{bychkovsky2011learning}, a dataset of RAW images captured by real SLR cameras.
These images already exhibit some lens blur from the capture process, and are further processed using DxO PhotoLab DeepPrime~\cite{dxo_deepprime}, which performs joint denoising and demosaicing but tends to smooth fine details as a side effect.
We select crops where the perceived blur is approximately spatially uniform.
We focus on this benchmark because most deblurring datasets primarily address motion blur or non-uniform defocus blur, while our method targets spatially uniform blur.
Since no ground truth sharp images are available, we provide only qualitative evaluation.

As shown in Fig.~\ref{fig:real_comparison}, ResShift and PASD fail to recover fine details, producing outputs that remain perceptually blurry.
SeeSR does not generate enough content, resulting in images with mostly flat surfaces.
DiffBIR generates sharper textures but introduces details that are not faithful to the input.
In contrast, our SupS models achieve both detail generation and fidelity to the input, producing more natural outputs. This is consistent with our observations on synthetic data.

\begin{figure}
  \setlength{\tabcolsep}{1pt} 
  \begin{tabular}{cccc}
    {\small{LQ}}& {\small{BM=0.7}} & {\small{BM=5.0}} & {\small{ BM=8.0}} \\
    \includegraphics[width=0.25\linewidth]{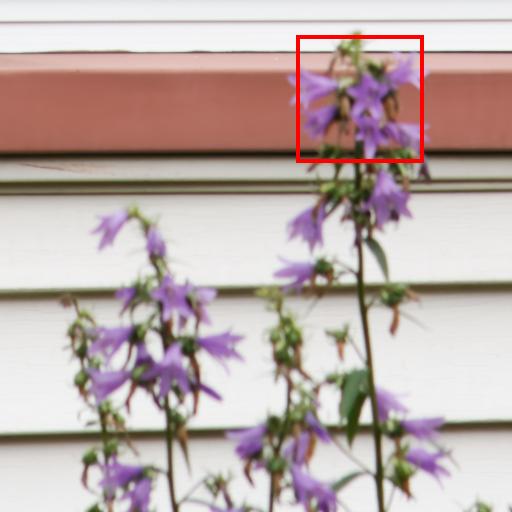} & 
    \includegraphics[width=0.25\linewidth]{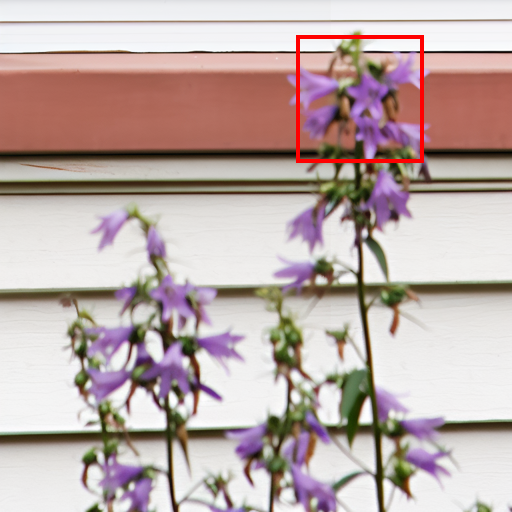} &
    \includegraphics[width=0.25\linewidth]{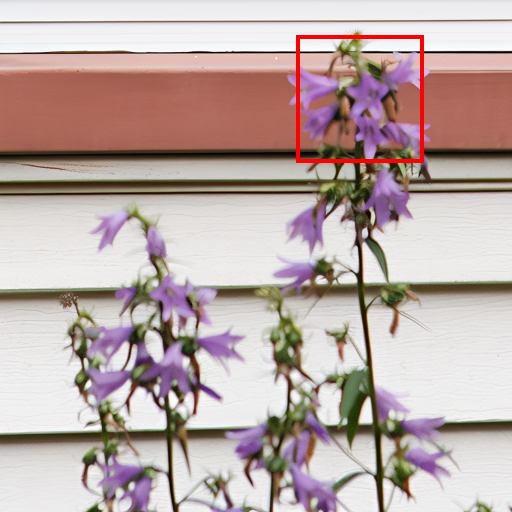} &
    \includegraphics[width=0.25\linewidth]{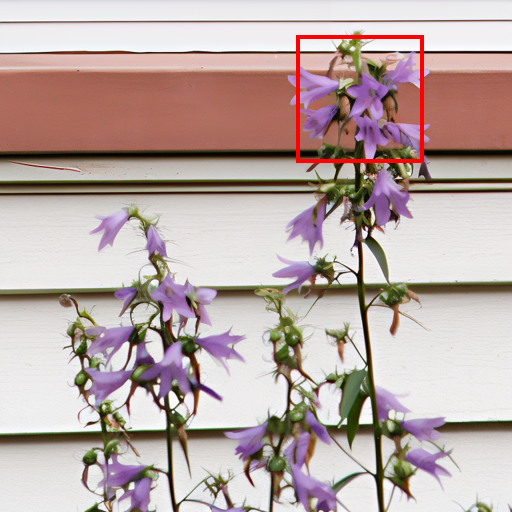} \\ [-0.3em]
    \includegraphics[width=0.25\linewidth]{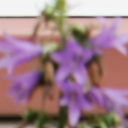} & 
    \includegraphics[width=0.25\linewidth]{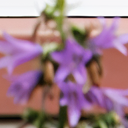} &
    \includegraphics[width=0.25\linewidth]{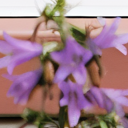} & 
    \includegraphics[width=0.25\linewidth]{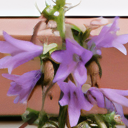} \\ [-0.1em]
  \end{tabular}
  \caption{Qualitative results on real images with different blur measure (BM) values. We see that higher BM values lead to stronger correction and sharper images.}
  \label{fig:real_img_results}
\end{figure}

On real images, we do not have access to the exact blur level, and therefore we manually adjust the blur measure $\blurm$ to control the generation strength.
As illustrated in Fig.~\ref{fig:real_img_results}, varying $\blurm$ effectively controls the amount of details generated, allowing users to adjust the balance between fidelity and generated content.
It is worth noting that the $\blurm$ value leading to perceptually optimal sharpness may not correspond to a physical blur kernel size.
\section{Conclusion}
\label{sec:conclusion}


We proposed SuperSharpen, a diffusion-based blind deblurring method with controllable restoration strength via a blur measure.
We compared ControlNet conditioning with full finetuning. Finetuning leads to better fidelity to the degraded input while ControlNet produces equally sharp but less faithful results.
Our method generalizes well to real-world blur from optical imperfections and RAW processing. 
However, it is restricted to spatially uniform blur, may produce artifacts on sensitive content such as text and faces, and does not yet preserve identity when the blur measure is close to zero.

\vfill\pagebreak

\bibliographystyle{IEEEbib}
\bibliography{refs}

\end{document}